\documentclass[10pt,conference, a4paper]{IEEEtran}
\usepackage{bm}
\usepackage{algorithm}
\usepackage[noend]{algpseudocode}
\usepackage{amsthm,amssymb,mathrsfs,setspace}
\usepackage{layout}
\usepackage{mathtools}
\usepackage{multirow}
\usepackage{enumitem}
\usepackage{amsmath}
\usepackage{amsthm}
\usepackage{url}
\usepackage{layout}
\usepackage[hidelinks]{hyperref}
\usepackage{epsfig}
\usepackage{xfrac}
\usepackage{booktabs}
\usepackage{epstopdf}
\usepackage{cite}
\usepackage{graphicx}

\usepackage{etoolbox}

\makeatletter
\patchcmd{\@makecaption}
  {\scshape}
  {}
  {}
  {}
\makeatother

\makeatletter
\def\footnoterule{\relax%
  \kern-5pt
  \hbox to \columnwidth{\hfill\vrule width 0.8\columnwidth height 0.4pt\hfill}
  \kern4.6pt}
\makeatother
\begin{document}

\title{Reinforcement Learning via Recurrent Convolutional Neural Networks}

\author{\IEEEauthorblockN{Tanmay Shankar}
\IEEEauthorblockA{Department of Mech. Engg. \\
IIT Guwahati\\
tanmay.shankar@gmail.com}
\and
\IEEEauthorblockN{Santosha K Dwivedy}
\IEEEauthorblockA{Department of Mech. Engg. \\
IIT Guwahati\\
dwivedy@iitg.ac.in}
\and 
\IEEEauthorblockN{Prithwijit Guha}
\IEEEauthorblockA{Department of EEE\\
IIT Guwahati\\
pguha@iitg.ac.in}}

\maketitle

\begin{abstract}

Deep Reinforcement Learning has enabled the learning of policies for complex tasks in partially observable environments, without explicitly learning the underlying model of the tasks. While such model-free methods achieve considerable performance, they often ignore the structure of task. We present a natural representation of to Reinforcement Learning (RL) problems using Recurrent Convolutional Neural Networks (RCNNs), to better exploit this inherent structure. We define 3 such RCNNs, whose forward passes execute an efficient Value Iteration, propagate beliefs of state in partially observable environments, and choose optimal actions respectively. Backpropagating gradients through these RCNNs allows the system to explicitly learn the Transition Model and Reward Function associated with the underlying MDP, serving as an elegant alternative to classical model-based RL. We evaluate the proposed algorithms in simulation, considering a robot planning problem. We demonstrate the capability of our framework to reduce the cost of re-planning, learn accurate MDP models, and finally re-plan with learnt models to achieve near-optimal policies. 

\end{abstract}
\IEEEpeerreviewmaketitle

\section{Introduction}
Deep Reinforcement Learning (DRL) algorithms exploit model-free Reinforcement Learning (RL) techniques, to achieve high levels of performance on a variety of tasks, often on par with human experts in the same domain \cite{mnih-atari-2013}. These DRL methods use deep networks to either approximate the action-value functions as in Deep Q Networks \cite{mnih-atari-2013, DBLP:journals/corr/HasseltGS15, DBLP:journals/corr/HausknechtS15}, or directly parametrizing the policy, as in policy gradient methods \cite{Sutton00policygradient}. While DRL methods adopt model-free approaches in order to generalize performance across various tasks, it is difficult to intuitively understand the reasoning of DRL approaches in making a particular choice of action, since they often ignore the underlying structure of the tasks.

\begin{figure*}
\includegraphics[width=\textwidth]{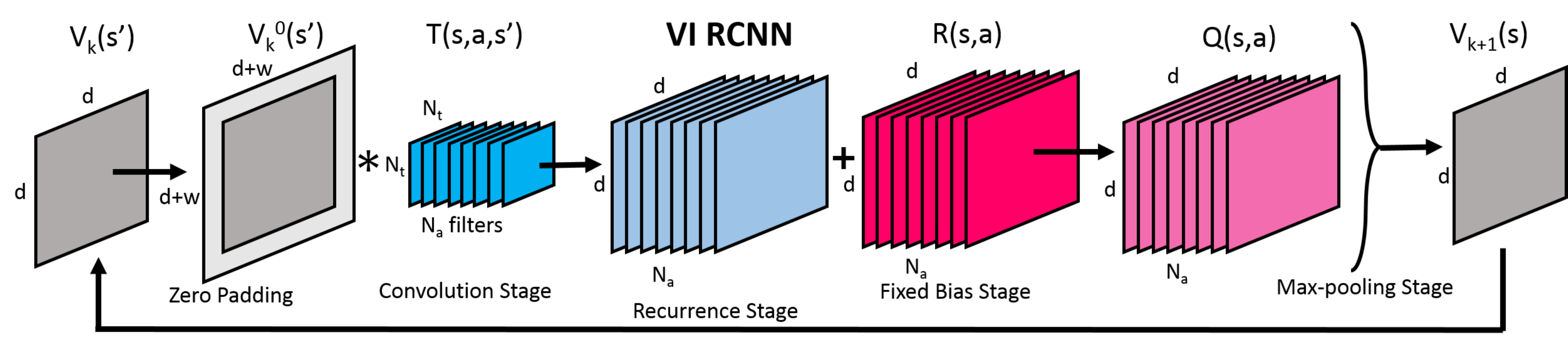}
\vspace{-2em}
\caption{\small{Schematic representation of the Value Iteration Recurrent Convolution Network. Notice the 4 stages present- convolution, Fixed-bias, Max-pooling and recurrence. The \textit{VI RCNN} facilitates a natural representation of the Bellman update as a single RCNN layer.}}
\vspace{-1em}
\label{fig:vi_rcnn}
\vspace*{-0.5em}
\end{figure*}

In contrast, model-based methods \cite{Atkeson+Santamaria:1997, DBLP:journals/ijrr/KoberBP13, Bakker:06icra, conf/icra/HesterQS10} exploit this inherent structure to make decisions, based on \textit{domain} knowledge. The estimates of the transition model and reward function associated with the underlying Markov Decision Process (MDP) are transferable across environments and agents \cite{Choi:2011:IRL:1953048.2021028}, and provide insight into the system's choice of actions. A significant deterrent from model-based RL is the indirect nature of learning with the estimated model; subsequent planning is required to obtain the optimal policy  \cite{Sutton:1998:IRL:551283}.

To jointly exploit this inherent structure and overcome this indirect nature, we present a novel fusion of Reinforcement and Deep Learning, by representing classical solutions to RL problems within the framework of Deep Learning architectures; in particular Recurrent Convolutional Neural Networks (RCNNs). By explicitly connecting the steps of solving MDPs with architectural elements of RCNNs, our representation inherits properties of such networks; allowing the use of backpropagation on the defined RCNNs as an elegant solution to model-based RL problems. The representation also exploits the inherent structure of the MDP to reduce the cost of replanning, further incentivizing model-based approaches.

The contributions of this paper are hence three-fold. We define a \textit{Value Iteration RCNN} (\textit{VI RCNN}), whose forward passes carry out Value Iteration to efficiently obtain the optimal policy. Second, we define a \textit{Belief Propagation RCNN} (\textit{BP RCNN}), to update beliefs of state via the Bayes Filter. Backpropagation through this network learns the underlying transition model of the agent in partially observable environments. Finally, we define a \textit{QMDP RCNN} by combining the \textit{VI RCNN} and the \textit{BP RCNN}. The \textit{QMDP RCNN} computes optimal choices of actions for a given belief of state. Backpropagation through the \textit{QMDP RCNN} learns the reward function of an expert agent, in a Learning from Demonstration via Inverse Reinforcement Learning (LfD-IRL) setting \cite{Abbeel04apprenticeshiplearning}. The learnt reward function and transition model may then be used to re-plan via the \textit{VI RCNN}, and the \textit{QMDP RCNN} to make optimal action choices for new beliefs and the learnt Q-values. \footnote{Please find our code and supplementary material at \ttfamily{ \url{https://github.com/tanmayshankar/RCNN_MDP}.}}


We note that \cite{DBLP:journals/corr/TamarLA16} follows an approach mathematically similar to the \textit{VI RCNN}; however it differs in being a model-free approach. The model-based approach followed here learns MDP models \textit{transferable} across environments and agents. Further, the \textit{QMDP RCNN} approach proposed here follows more naturally than the fully connected layer used in \cite{DBLP:journals/corr/TamarLA16}. The gradient updates of the \textit{QMDP RCNN} adopt an intuitive form, that further contributes to an intuitive understanding of the action choices of the \textit{QMDP RCNN}, as compared to \cite{DBLP:journals/corr/TamarLA16}.


We evaluate each of the RCNNs proposed in simulation, where the given task is a 2-D robot planning problem. We demonstrate the capacity of the \textit{VI RCNN} to reduce the cost of re-planning by significantly reducing the execution time as compared to classical Value Iteration. We evaluate the \textit{BP RCNN} based on the accuracy of the learnt transition models against ground truth models, and show that it appreciably outperforms naive model-based methods for partially observable environments. Our experiments finally demonstrate that the \textit{QMDP RCNN} is able to generate policies via re-planning with learnt MDP models, that accurately represent policies generated from ground truth MDP models.


\vspace*{-0.2em}
\section{The Value Iteration RCNN}
In this section, we formulate Value Iteration as a recurrent convolution, and hence present the \textit{Value Iteration RCNN} (\textit{VI RCNN}) to carry out Value Iteration. We consider a standard Markov Decision Process, consisting of a 2-dimensional state space $S$ of size $N_d \times N_d$, state $s \in S$, action space $A$ of size $N_a$, actions $a \in A$, a transition model $T(s,a,s')$, reward function $R(s,a)$, and discount factor $\gamma$. Value Iteration, which is typically used in solving for optimal policies in an MDP, invokes the Bellman Update equation as $V_{k+1}(s) =  \max_a \ R(s,a) + \gamma \sum_{s'} T(s,a,s') V_{k} (s')$.

\noindent \textit{\textbf{Value Iteration as a Convolution:}}
In our 2-D state space, $S$, states $s$ and $s'$ are defined by 2 pairs of indices $(i,j)$ and $(p,q)$. The summation over all states $s'$ may hence be equivalently represented as $\sum_{s'} = \sum_{p=1}^{N_d} \sum_{q=1}^{N_d}$. The transition model is hence of size $N_d^2 \times N_a \times N_d^2$. We may express $V_{k+1}(s)$ as $\max_a R(s,a) + \gamma \sum_{p=1}^{N_d} \sum_{q=1}^{N_d} T(s,a,s')_{p,q} V_{k}(s')_{p,q}$. 

Actions correspond to transitions in the agent's state, dictated by the internal dynamics of the agent, or in the case of physical robots, restricted by the robot's configuration and actuator capacities. It is impossible for the agent to immediately move from its current state to a far away state. This allows us to define a $w$ neighborhood centred around the state $s'$, $W(s')$, so the agent only has a finite probability of transitioning to other states within $W$. Mathematically, $T(s,a,s') \neq 0$ if $|i-p| \leq w$ and $|j-q| \leq w$ or if $s \in W(s')$, and $T(s,a,s') = 0$ for $s \not\in W(s')$. 

Further, the transition model is often invariant to the spatial location of the occurring transition. For example, a differential drive robot's dynamics are independent of the robot's position. We assume the transition model is stationary over the state space, incorporating appropriate state boundary conditions. We may now visualize this $w$ neighborhood restricted transition model itself to be centred around $s'$. By defining a \textit{flipped transition model} as: $\overline {T}(s,a,s')_{m,n}= \ T(s,a,s')_{N_t - m, N_t - n}$, and indexing it by $(u,v)$, we may express $V_{k+1}(s)$ as $\max_a R(s,a) + \gamma \sum_{u=-w}^{w} \sum_{v=-w}^{w} \overline{T} (s,a,s')_{u,v} V_{k}(s')_{i-u,j-v} $.
\noindent
We may represent this as a convolution:
\begin{equation}
V_{k+1}(s) =  \max_a \ R(s,a) + \gamma \ \overline{T}(s,a,s') * V_{k}(s')
\label{conv}
\end{equation}

\noindent \textit{\textbf{The Value Iteration RCNN:}}
We define a \textit{Value Iteration RCNN} (\textit{VI RCNN}) to represent classical Value Iteration, based on the correlation between the Bellman update equation, and the architectural elements of an RCNN. Equation \eqref{conv} can be thought of as a single layer of a recurrent convolutional neural network consisting of the following 4 stages:
\begin{enumerate}[leftmargin=*]
	\item \textit{Convolutional Stage:} The convolution of $\overline{T}(s,a,s') * V_{k}(s')$ represents the convolutional stage. 
	\item \textit{Max-Pooling Stage:} The maximum at every state $s$ taken over all actions $a$, $\max_a$ is analogous to a max-pooling stage along the action `channel'. 
	\item \textit{Fixed Bias Stage:} The introduction of the reward term $R(s,a)$ is treated as addition of a fixed bias. 
	\item \textit{Recurrence Stage:} As $k$ is incremented in successive iterations, $V_{k+1}(s)$ is fed back as an `input' to the network, introducing a recurrence relation into the network. 
\end{enumerate}

We may think of $V_k(s')$ as a single-channel $N_d \times N_d$ image. $\overline{T}(s,a,s')$ then corresponds to a series of $N_a$ \textit{transition filters}, each of $N_t \times N_t$ size ($N_t=2w+1$), each to be convolved with the image. The values of the transition filters correspond directly to transition probabilities between states $s$ and $s'$, upon taking action $a$. Note that these convolutional \textit{transition filters} naturally capture the spatial invariance of transitions by virtue of lateral parameter sharing inherent to convolutional networks. Further, in this representation, each \textit{transition filter} corresponds directly to a particular action. This unique one-to-one mapping between actions and filters proves to be very useful in learning MDP or RL models, as we demonstrate in section III. Finally, the \textit{VI RCNN} is completely differentiable, and can be \textit{trained} by applying backpropagation through it.

\section{The Belief Propagation RCNN}
In this section, we present the \textit{Belief Propagation RCNN} (\textit{BP RCNN}), to represent the Bayes filter belief update within the architecture of an RCNN, and hence learn MDP transition models. For an agent to make optimal choices of actions in partially observable environments, an accurate belief of state $b(s)$ is required. 
Upon taking an action $a$ and receiving observation $z$, we may use the Bayes filter to propagate the belief $b(s')$ in terms of an observation model $O(s',a,z)$, associated with probability $p(z|s',a)$. The discrete Bayes filter is depicted in \eqref{bayes_filter_eqn}. Note the denominator is equivalent to $p(z|a,b)$, and can be implemented as a normalization factor $\eta$.
\begin{equation}
b(s') =  \frac {O(s',a,z) \sum_{s \in S} T(s,a,s') b(s)} {\sum_{s' \in S} O(s',a,z) \sum_{s \in S} T(s,a,s') b(s)} 
\label{bayes_filter_eqn}
\end{equation}

\begin{figure}
\includegraphics[width=0.5\textwidth]{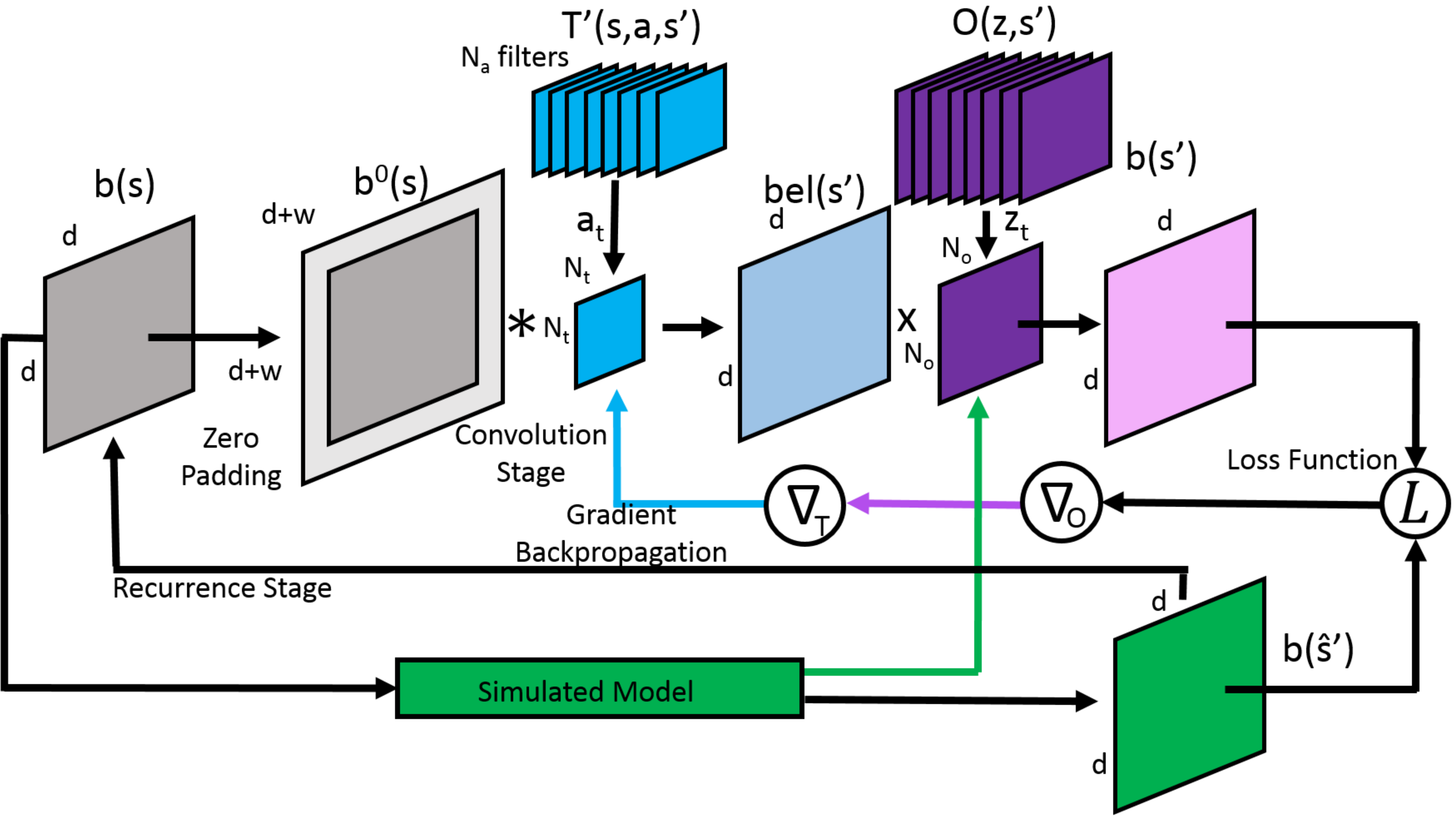}
\vspace{-2em}
\caption{\small{Schematic representation of the Belief Propagation Recurrent Convolution Network. Forward passes represented by black arrows, while backpropagation is symbolized by blue arrows.}}
\vspace{-1em}
\label{fig:bp_rcnn}
\vspace*{-0.5em}
\end{figure}

\noindent \textit{\textbf{Bayes Filter as a Convolution and an Element-wise Product:}}
We may represent the motion update of the traditional Bayes filter as a convolution, analogous to the representation of $\sum_{s'} T(s,a,s') V_{k}(s')$ step of Value Iteration as $\overline{T}$  $(s,a,s')*V_k(s')$ in equation \eqref{conv}. Transitions in a Bayes filter occur from $s$ to $s'$, in contrast to the `backward' passes executed in Value Iteration, from $s'$ to $s$. This suggests our transition model is centred around $s$ rather than $s'$, and we may represent $\sum_{s}$ as $\sum_{i=1}^{N_d} \sum_{j=1}^{N_d}$. We write the intermediate belief $\overline{b(s')}_{p,q}$ as $\sum_{u=-w}^{w} \sum_{v=-w}^{w} T(s,a,s')_{u,v} \ b(s)_{p-u,q-v}$, which can be expressed as a convolution $\overline{b(s')} = T(s,a,s') * b(s)$.
For the Bayes Filter correction update, we consider a simple observation model $O(z,s')$ for observation $z$ in state $s'$. The agent observes its own state as $s^{z}_{x,y}$, indexed by $x$ and $y$, given that the actual state of the robot is $s'_{i,j}$, with a probability $p(s^{z}_{x,y} | s'_{i,j})$. Analogous to the $w$ neighborhood for the transition model, we introduce a $h$ neighborhood for the observation model, centred around the observed state $s^z_{x,y}$. Thus $O(s^z_{x,y},s') \neq 0$ for states $s'_{i,j}$ within the $h$ neighborhood $H(s^z_{x,y})$, and $O(s^z_{x,y},s')=0$ for all $s'_{i,j} \not\in H(s^z_{x,y})$. 

The correction update in the Bayes filter can be represented as a masking operation on the intermediate belief $\overline{b(s')}$, or an element-wise (Hadamard) product ($\odot$). Since the observation $O(s^z)$ is also centred around $s^z$, we may express $b(s')_{i,j}$ as $\eta \ O(s^z)_{i-x,j-y} \ \overline{b(s')}_{i,j} \ \forall \ i, j $, which corresponds to a normalized element-wise product $b(s') = \eta \ O(s^z) \odot \overline{b(s')}$.
Upon combining the motion update convolution and the correction update element-wise product, we may represent the Bayes Filter as a convolutional stage followed by an element-wise product. We have:
\begin{equation}
b(s') = \eta \ O(s^z) \odot T(s,a,s') * b(s) 
\label{belief_prop_rcnn}
\end{equation}

\noindent \textit{\textbf{The Belief Propagation RCNN:}}
We define a \textit{Belief Propagation RCNN} (\textit{BP RCNN}) to represent the Bayes Filter belief update, analogous to the representation of Value Iteration as the \textit{VI RCNN}. A single layer of the \textit{BP RCNN} represents equation \eqref{belief_prop_rcnn}, consisting of the following stages: 
\begin{enumerate}[leftmargin=*]
	\item \textit{Convolution Stage}: The convolution $T(s,a,s') * b(s)$ represents the convolution stage of the \textit{BP RCNN}. 
	\item \textit{Element-wise Product Stage}: The element-wise multiplication $O(s^z) \odot \overline{b(s')}$, can be considered as an element-wise product (or Hadamard product) layer in 2D. 
	\item \textit{Recurrence Stage}: The network output, $b(s')$, is fed as the input $b(s)$ at the next time step, forming an \textit{output recurrence} network.
\end{enumerate}

Forward passes of the \textit{BP RCNN} propagate beliefs of state, given a choice of action $a_t$ and a received observation $z_t$. The belief of state at any given instant, $b(s)$, is treated as a single channel $N_d \times N_d$ image, and is convolved with a \textit{single} transition filter $T_{a_t}$, corresponding to the action executed, $a_t$. A key difference between the \textit{VI RCNN} and the \textit{BP RCNN} is that only a single transition filter is \textit{activated} during the belief propagation, since the agent may only execute a single action at any given time instant. The \textit{BP RCNN} is also completely differentiable. 

\noindent \textit{\textbf{Training and Loss:}}
Learning the weights of the \textit{BP RCNN} via backpropagation learns the transition model of the agent. Our objective is thus to learn a transition model $T'(s,a,s')$ such that network output $b(s')$ is as close to the target belief $\widehat{b(s')}$ as possible, at every time step. Formally, we minimize a loss function defined as the least square error between both beliefs over all states: 
$L_t = \sum_{i=1}^{N_d} \sum_{j=1}^{N_d} \ \big( \widehat{b(s'_t)} - b(s'_t) \big) _{i,j}^2$,
with conditions on the transition model,  $0 \leq T'(s,a,s')_{m,n} \leq 1$, and $\sum_{m=-w}^{w} \sum_{n=-w}^w T'(s,a,s')_{m,n}=1 \ \forall \ a \in A$, being incorporated using appropriate penalties. 

The \textit{BP RCNN} is trained in a typical RL setting, with an agent interacting with an environment by taking random action choices $(a_t,a_{t+1}...)$, and receiving corresponding observations, $(z_t,z_{t+1}...)$. The target beliefs, $\widehat{b(s')}$, are generated as one-hot representations of the observed state. While training the \textit{BP RCNN}, the randomly initialized transition model magnifies the initial uncertainty in the belief as it is propagated forward in time, leading to instability in learning the transition model. Teacher forcing \cite{Goodfellow-et-al-2016-Book} is adopted to handle this uncertainty; thus the \textit{target belief}, rather than the network output, is propagated as the input for the next time step. Since the target belief is independent of the initial uncertainty of the transition model, the network is able to learn a meaningful set of filter values. Teacher forcing, or such \textit{target recurrence}, also decouples consecutive time steps, reducing the \textit{backpropagation through time} to backpropagation over data points that are only generated sequentially. 

\begin{figure*}[!t]
\includegraphics[width=\textwidth]{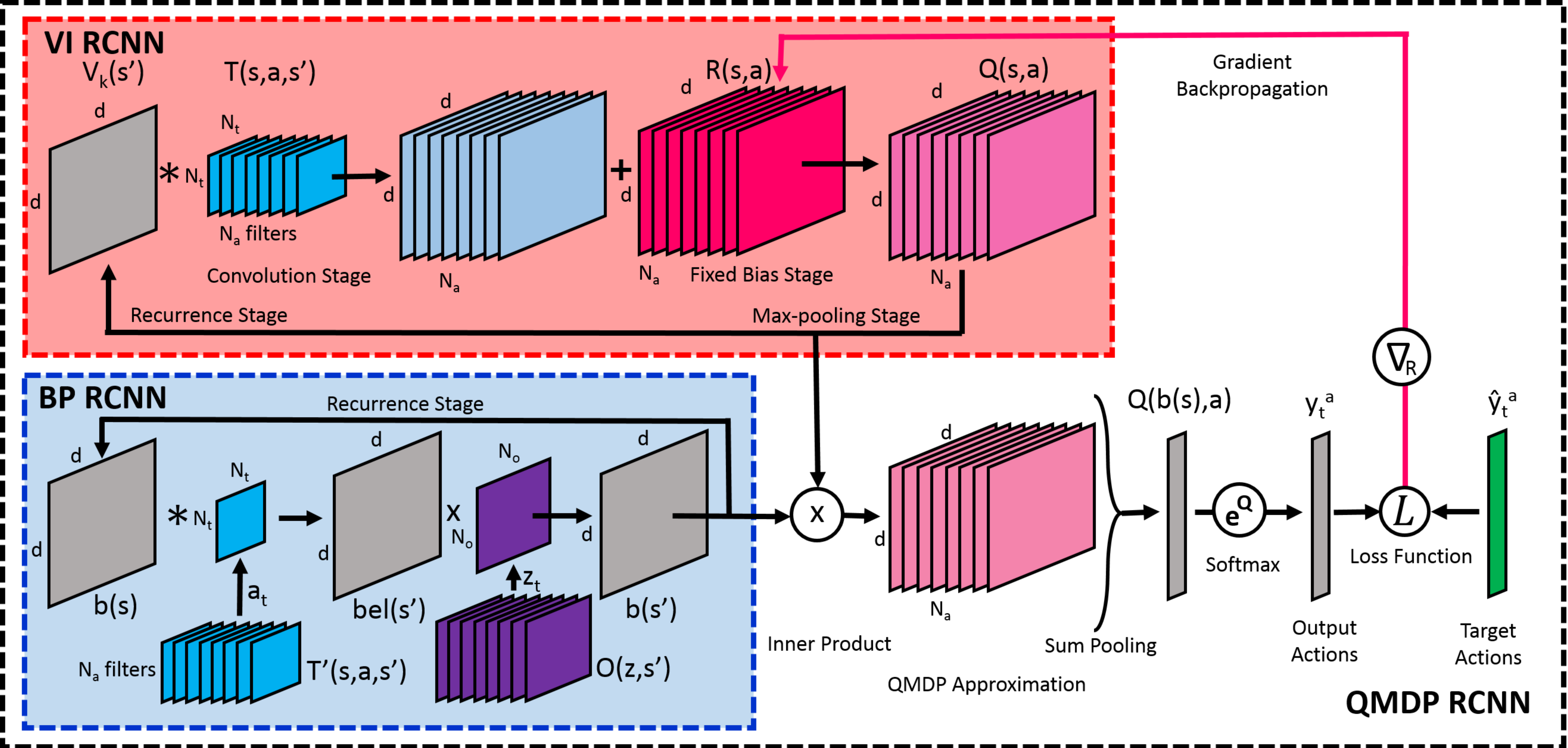}
\vspace{-2em}
\caption{\small{Schematic representation of the \textit{QMDP RCNN}, as a combination of the Value Iteration RCNN and the Belief Propagation RCNN. Notice the following stages: \textit{inner product, sum-pooling}, and the \textit{softmax} stage providing the output actions. The gradients propagated to the reward function may alternately be propagated to the Q-value estimate. Forward passes of the \textit{QMDP RCNN} are used for planning in partially observable domains, and backpropagation learns the reward function of the MDP. The \textit{BP RCNN} component can be trained in parallel to the \textit{QMDP RCNN}.}}
\vspace{-1em}
\label{fig:qmdp_rcnn}
\end{figure*}

\section{The QMDP RCNN}
We finally present the \textit{QMDP RCNN} as a combination of the \textit{VI RCNN} and the \textit{BP RCNN}, to retrieve optimal choices of actions, and learn the underlying reward function of the POMDP (in addition to the transition model).
Treating the outputs of the \textit{VI RCNN} and the \textit{BP RCNN} as the action-value function, $Q(s,a)$, and the belief of state, $b(s)$ respectively at every time step, we may fuse these outputs to compute belief space Q-values $Q(b(s),a)$ using the QMDP approximation. We have $Q(b(s),a) = \sum_{s \in S} Q(s,a)_{\tiny{MDP}} b(s)$, which may alternately be represented as a Frobenius Inner product ${\langle b(s), Q(s,a) \rangle}_F$. 
Optimal action choices $\mathbf{y}^a_t$, corresponding to the highest belief space Q-values, and can be retrieved from the softmax of $Q(b(s),a)$: 
\begin{equation}
\mathbf{y}^a_t = \frac{e^{Q(b(s),a)}}{\sum_a e^{Q(b(s),a)}}
\end{equation}

\begin{table*}[!t]
\centering
\caption{Transition Model Error and Replanning Policy Accuracy for \textit{BP RCNN}.}
\vspace*{-0.9em}
\scriptsize{Values in Bold represent best-in-class performance. Parameter values marked * are maintained during variation of alternate parameters, and are used during final experiments.}
\label{trans_error_table}
\renewcommand{\arraystretch}{1.2}
\resizebox{\linewidth}{!}{
\begin{tabular}{|c|c|c|c|c|c|c|c|}
\hline
 & Algorithm: & \multicolumn{2}{c|}{Belief Propagation RCNN} & \multicolumn{2}{c|}{Weighted Counting} & \multicolumn{2}{c|}{Naive Counting} \\ \hline
Parameter & Type & Transition Error & Replanning Accuracy & Transition Error & Replanning Accuracy & Transition Error & Replanning Accuracy \\ \hline
\multirow{2}{*}{\begin{tabular}[c]{@{}c@{}}Environment\\ Observability\end{tabular}} & Fully Observable & \textbf{0.0210} & \textbf{96.882 \%} & 0.0249 & 96.564 \% & 0.0249 & 96.564 \% \\ \cline{2-8} 
 & Partially Observable* & 0.1133 & 95.620 \% & 1.0637 & 75.571 \% & 0.1840 & 83.590 \% \\ \hline
\multirow{2}{*}{Teacher Forcing} & Output Recurrence & 3.5614 & 21.113 \% & 3.0831 & 27.239 \% & \multicolumn{2}{c|}{\multirow{2}{*}{No use of Recurrence.}} \\ \cline{2-6}
 & Target Recurrence* & \textbf{0.1133} & \textbf{95.620 \%} & 1.0637 & 75.571 \% & \multicolumn{2}{c|}{} \\ \hline
\multirow{3}{*}{\begin{tabular}[c]{@{}c@{}}Learning Rate\\ Adaptation\end{tabular}} & RMSProp & 2.8811 & 40.315 \% & 2.5703 & 44.092 \% & \multicolumn{2}{c|}{\multirow{3}{*}{No use of Learning Rate.}} \\ \cline{2-6}
 & Linear Decay & 0.2418 & 93.596 \% & 1.3236 & 52.451 \% & \multicolumn{2}{c|}{} \\ \cline{2-6}
 & Filter-wise Decay* & \textbf{0.1133} & \textbf{95.620 \%} & 1.0637 & 75.571 \% & \multicolumn{2}{c|}{} \\ \hline
\end{tabular}}
\vspace*{-1.4em}
\end{table*}

\noindent \textit{\textbf{The QMDP RCNN:}}
In addition to the stages of the \textit{VI RCNN} and \textit{BP RCNN}, the \textit{QMDP RCNN} is constructed from the following stages:
\vspace*{-0.05in}
\begin{enumerate}[leftmargin=*]
	\item \textit{Frobenius Inner Product Stage:} The QMDP approximation step $Q(b(s),a) = {\langle b(s) , Q(s,a) \rangle}_F $. represents an Frobenius inner product stage. This can be implemented as a \textit{`valid'} convolutional layer with \textit{zero} stride.

	\item \textit{Softmax Stage:} $\mathbf{y}^a_t = softmax \ Q(b(s),a)$ represents the softmax stage of the network.
\end{enumerate}

The resultant \textit{QMDP RCNN}, as depicted in Figure \ref{fig:qmdp_rcnn}, combines planning and learning in a single network. Forward passes of the network provide action choices $\mathbf{y}^a_t$ as an output, and backpropagating through the network learns the reward function, and updates Q-values and optimal action choices via planning through the \textit{VI RCNN} component. 

\noindent \textit{\textbf{Training and Loss:}}
The \textit{QMDP RCNN} is trained in a Learning from Demonstration - Inverse Reinforcement Learning setting. An expert agent demonstrates a set of tasks, recorded as a series of trajectories $\{(a_t, z_t), (a_{t+1},z_{t+1})...\}$, which serve as inputs to the network. The actions executed in the expert trajectories are represented in a one-hot encoding, serving as the network targets $\widehat{\mathbf{y}}^a_t$. The objective is to hence learn a reward function such that the action choices output by the network match the target actions at every time step. 
The loss function chosen is the cross-entropy loss between the output actions and the target actions at a given time step, defined as $C_t = - \sum_a {\widehat{\mathbf{y}}^a_t ln \mathbf{y}^a_t}$. The \textit{QMDP RCNN} learns the reward function by backpropagating the gradients retrieved from this loss function through the network, to update the reward function. The updated reward estimate is then fed back to the \textit{VI RCNN} component to update the Q-values, and hence the action choices of the network. Experience replay \cite{Lin:1992} is used to randomize over the expert trajectories and transitions while training the \textit{QMDP RCNN}. 

The closed form of the gradient for reward updates is of the form of $R(s,a) \xleftarrow{\alpha_t} -(\mathbf{y}^a_t - \widehat{\mathbf{y}}^a_t) b(s)$. The $(\mathbf{y}^a_t - \widehat{\mathbf{y}}^a_t)$ term thus dictates the \textit{extent to which} actions are positively or negatively \textit{reinforced}, while the belief term $b(s)$ acts in a manner similar to an \textit{attention mechanism}, which directs the reward function updates to specific regions of the state space where the agent is believed to be. 

We emphasize that the \textit{QMDP RCNN} differs from traditional DRL approaches in that the \textit{QMDP RCNN} is not provided with samples of the reward function itself via its interaction with the environment, rather, it is provided with action choices made by the expert. While the LfD-IRL approach employed for the \textit{QMDP RCNN} is on-policy, the in-built planning (via the \textit{VI RCNN}) causes reward updates to permeate the entire state space. The dependence of the LfD-IRL approach in the \textit{QMDP RCNN} on expert-optimal actions for training differs from the \textit{BP RCNN}, which uses arbitrary (and often suboptimal) choices of actions.


\section{Experimental Results and Discussions}
In this section, experimental results are individually presented with respect to each of the 3 RCNNs defined. 

\begin{figure}[!t]
\includegraphics[width=0.5\textwidth]{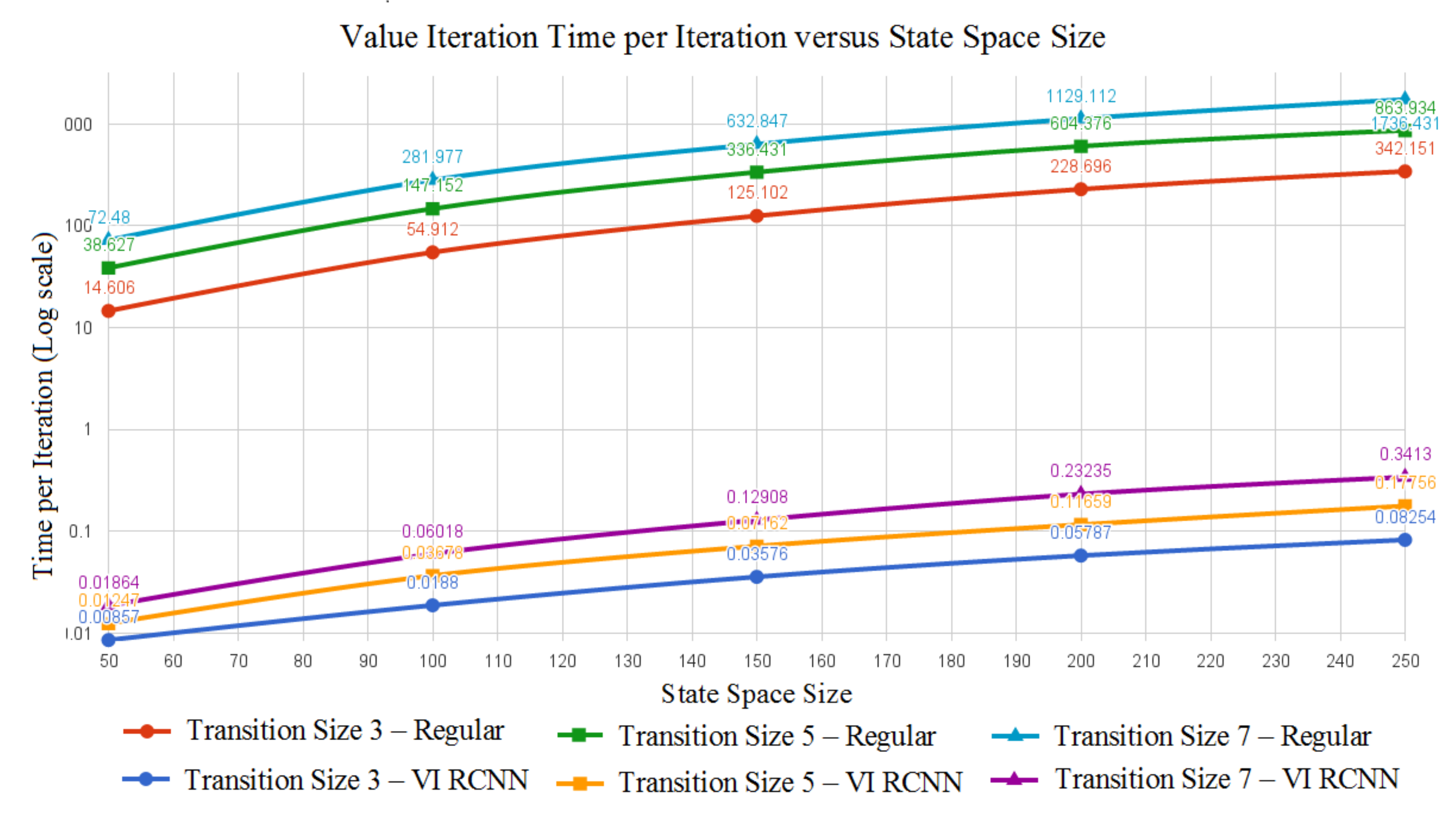}
\vspace{-2em}
\caption{\small{Plot of the time per iteration of Value Iteration versus size of the State Space. The time taken by the \textit{VI RCNN} is orders of magnitude lesser than regular Value Iteration, for all transition space sizes.}}
\vspace{-1em}
\label{fig:log_time}

\end{figure}

\noindent \textit{\textbf{VI RCNN:}} Here, our contribution is a representation that enables a more efficient computation of Value Iteration. We thus present the per-iteration run-time of Value Iteration via the \textit{VI RCNN}, versus a standard implementation of Value Iteration. Both algorithms are implemented in Python, run on an Intel Core i7 machine, 2.13 GHz, with 4-cores. The inherent parallelization of the convolution facilitates the speed-up of VI by several orders of magnitude, as depicted in Figure \ref{fig:log_time}; at best, the \textit{VI RCNN} completes a single iteration $5106.42$ times faster, and at worst, it is $1704.43$ times faster than the regular implementation.

\noindent\textbf{\textit{BP RCNN:}} The primary objective while training the \textit{BP RCNN} is to determine an accurate estimate of the transition model. We evaluate performance using the least square error between the transition model estimate and the ground truth model, defined as $C'_t = \sum_a \sum_u \sum_v {(T^a_{u,a} - T'^a_{u,v})}^2 $. Since the final objective is to use these learnt models to generate new policies by replanning via the \textit{VI RCNN}, 
we also present the replanning accuracy of each model; defined as the percentage of actions chosen correctly by the network's policy, compared to the original policy. A comparison is provided against the models learnt using a counting style algorithm analogous to that used in \cite{Kearns:2002:NRL:599616.599699}, as well as a weighted-counting style algorithm that updates model-estimates by counting over belief values. We experiment with the performance of these algorithms under both fully and partially observable settings, and whether teacher forcing is used (via target recurrence) or not (as in output recurrence). Different methods of adapting the learning rate are explored, including RMSProp, linear decay, and maintaining individual learning rates for each transition filter, or \textit{filter-wise decay}, as presented in Table \ref{trans_error_table}. 

\begin{table*}[!t]
\centering
\caption{Replanning Accuracy and Increase in Expected Reward for Inverse Reinforcement Learning from Demonstrations via the \textit{QMDP RCNN}.}
\vspace*{-0.8em}
\scriptsize{Values in Bold represent best-in-class performance. Parameter values marked * are maintained during variation of alternate parameters, and are used during final experiments.}
\label{table_3}
\renewcommand{\arraystretch}{1.2}
\tiny
\resizebox{\linewidth}{!}{
\begin{tabular}{|c|c|c|l|c|c|c|c|c|c|c|}
\hline
\multicolumn{2}{|c|}{Algorithm:} & \multicolumn{7}{c|}{QMDP RCNN} & \multirow{3}{*}{\begin{tabular}[c]{@{}c@{}}Weighted\\ Counting\end{tabular}} & \multirow{3}{*}{\begin{tabular}[c]{@{}c@{}}Naive\\ Counting\end{tabular}} \\ \cline{1-9}
\multicolumn{2}{|c|}{Parameter:} & \multicolumn{3}{c|}{Learning Rate} & \multicolumn{2}{c|}{Experience Replay} & \multicolumn{2}{c|}{Feedback Recurrence} &  &  \\ \cline{1-9}
Environment: & Parameter Value: & \multicolumn{2}{c|}{RMSProp} & Linear Decay* & None & Used* & Immediate & Delayed* &  &  \\ \hline
\multirow{2}{*}{\begin{tabular}[c]{@{}c@{}}Learnt Reward\\ Learnt Transition\end{tabular}} & Replanning Accuracy & \multicolumn{2}{c|}{45.113 \%} & \textbf{65.120 \%} & 62.211 \% & \textbf{65.120 \%} & 65.434 \% & \textbf{65.120 \%} & 45.102 \% & 50.192 \% \\ \cline{2-11} 
 & Expected Reward Increase & \multicolumn{2}{c|}{-75.030 \%} & \textbf{-10.688 \%} & -16.358 \% & \textbf{-10.688 \%} & -16.063 \% & \textbf{-10.688 \%} & -74.030 \% & -43.166 \% \\ \hline
\multirow{2}{*}{\begin{tabular}[c]{@{}c@{}}Learnt Reward\\ Known Transition\end{tabular}} & Replanning Accuracy & \multicolumn{2}{c|}{43.932 \%} & \textbf{63.476 \%} & 60.852 \% & \textbf{63.476 \%} & 63.900 \% & \textbf{63.476 \%} & 49.192 \% & 53.427 \% \\ \cline{2-11} 
 & Expected Reward Increase & \multicolumn{2}{c|}{-79.601 \%} & \textbf{-11.429 \%} & -16.106 \% & \textbf{-11.429 \%} & -17.949 \% & \textbf{-11.429 \%} & -64.294 \% & -31.812 \% \\ \hline
\multirow{2}{*}{\begin{tabular}[c]{@{}c@{}}Known Reward\\ Learnt Transition\end{tabular}} & Replanning Accuracy & \multicolumn{7}{c|}{\textbf{95.620 \%} (Known Reward used for Replanning)} & 75.571 \% & 83.590 \% \\ \cline{2-11} 
 & Expected Reward Increase & \multicolumn{7}{c|}{\textbf{+0.709 \%} (Known Reward used for Replanning)} & -0.617 \% & -0.806 \% \\ \hline
\end{tabular}}
\vspace*{-1em}
\end{table*}

The \textit{BP RCNN's} loss function is defined in terms of the non-stationary beliefs at any given time instant. An online training approach is hence rather than batch mode. We observe the \textit{filter-wise decay} outperforms linear decay when different actions are chosen with varying frequencies. RMSProp suffers from the oscillations in the loss function that arise due to this dynamic nature, and hence performs poorly. The use of teacher forcing mitigates this dynamic nature over time, increasing replanning accuracy from $21.11 \%$ to $95.62 \%$. The $95.62 \%$ replanning accuracy attained under partial-observability approaches the $96.88 \%$ accuracy of the algorithm under fully observable settings.

\noindent\textit{\textbf{QMDP RCNN:}} For the \textit{QMDP RCNN}, the objective is learn a reward function that results in a similar policy and level of performance as the original reward. Since learning such a reward such that the demonstrated trajectories are optimal does not have a unique solution, quantifying the accuracy of the reward estimate is meaningless. Rather, we present the replanning accuracy (as used in the \textit{BP RCNN}) for both learnt and known transition models and rewards. We also run policy evaluation on the generated policy using the original rewards, and present the increase in expected reward for the learnt models; defined as $\frac{1}{N_d^2} \sum_s \frac{V(s)_{learnt} - V(s)_{orig}} {V(s)_{orig}}$ . The results are presented in Table \ref{table_3}. 

As is the case in the \textit{BP RCNN}, RMSProp (and adaptive learning rates in general) counter the magnitude of reward updates dictated by the \textit{QMDP RCNN}, and hence adversely affect performance. Experience Replay marginally increases the replanning accuracy, but has a significant effect on the increase in expected reward. Similarly, using delayed feedback (after passes through an entire trajectory) also boosts the increase in expected reward. On learning both rewards and transition models, the \textit{QMDP RCNN} achieves an appreciable $65.120 \%$ policy error and a minimal $-10.688 \%$ change in expected reward. We emphasize that given access solely to observations and action choices of the agent, and without assuming any feature engineering of the reward, the \textit{QMDP RCNN} is able to achieve near-optimal levels of expected reward. Utilizing the ground truth transition with the learnt reward, the \textit{QMDP RCNN} 
performs marginally worse, with a $63.476 \%$ accuracy and a $-11.429 \%$ change in expected reward. 

The true efficacy of the \textit{BP RCNN} and the \textit{QMDP RCNN} lie in their ability to learn accurate transition models and reward functions in partially observable settings, where they outperforms existing model-based approaches of naive and weighted counting by significant margins, in terms of replanning accuracy, transition errors, and expected reward. Finally, we note that while all $3$ defined RCNN architectures are demonstrated for 2-D cases, these architectures can be extended to any number of dimensions, and number of actions, with suitable modifications to the convolution operation in higher dimensions.

\vspace*{-0.5em}
\section{Conclusion}

In this paper, we defined $3$ RCNN like architectures, namely the \textit{Value Iteration RCNN}, the \textit{Belief Propagation RCNN}, and the \textit{QMDP RCNN}, to facilitate a more natural representation of solutions to model-based Reinforcement Learning. Together, these contributions speed up the planning process in a partially observable environment, reducing the cost of replanning for model-based approaches. Given access to agent observations and action choices over time, the \textit{BP RCNN} learns the transition model, and the \textit{QMDP RCNN} learns the reward function, and subsequently replans with learnt models to make near-optimal action choices. The proposed architectures were also found to outperform existing model-based approaches in speed and model accuracy. The natural symbiotic representation of planning and learning algorithms allows these approaches to be extended to more complex tasks, by integrating them with sophisticated perception modules.
\vspace*{-0.5em}
\nocite{*}

\bibliographystyle{IEEEtran}
\bibliography{/home/tanmay/Research/RCNN_MDP/ICPR/New_Draft/REV_1/Paper_10/ICPR_REF.bib}

\begin{thebibliography}{10}
\providecommand{\url}[1]{#1}
\csname url@samestyle\endcsname
\providecommand{\newblock}{\relax}
\providecommand{\bibinfo}[2]{#2}
\providecommand{\BIBentrySTDinterwordspacing}{\spaceskip=0pt\relax}
\providecommand{\BIBentryALTinterwordstretchfactor}{4}
\providecommand{\BIBentryALTinterwordspacing}{\spaceskip=\fontdimen2\font plus
\BIBentryALTinterwordstretchfactor\fontdimen3\font minus
  \fontdimen4\font\relax}
\providecommand{\BIBforeignlanguage}[2]{{%
\expandafter\ifx\csname l@#1\endcsname\relax
\typeout{** WARNING: IEEEtran.bst: No hyphenation pattern has been}%
\typeout{** loaded for the language `#1'. Using the pattern for}%
\typeout{** the default language instead.}%
\else
\language=\csname l@#1\endcsname
\fi
#2}}
\providecommand{\BIBdecl}{\relax}
\BIBdecl

\bibitem{mnih-atari-2013}
V.~Mnih, K.~Kavukcuoglu, D.~Silver, A.~Graves, I.~Antonoglou, D.~Wierstra, and
  M.~Riedmiller, ``Playing atari with deep reinforcement learning,'' in
  \emph{NIPS Deep Learning Workshop}, 2013.

\bibitem{DBLP:journals/corr/HasseltGS15}
\BIBentryALTinterwordspacing
H.~van Hasselt, A.~Guez, and D.~Silver, ``Deep reinforcement learning with
  double q-learning,'' \emph{CoRR}, vol. abs/1509.06461, 2015. [Online].
  Available: \url{http://arxiv.org/abs/1509.06461}
\BIBentrySTDinterwordspacing

\bibitem{DBLP:journals/corr/HausknechtS15}
\BIBentryALTinterwordspacing
M.~J. Hausknecht and P.~Stone, ``Deep recurrent q-learning for partially
  observable mdps,'' \emph{CoRR}, vol. abs/1507.06527, 2015. [Online].
  Available: \url{http://arxiv.org/abs/1507.06527}
\BIBentrySTDinterwordspacing

\bibitem{Sutton00policygradient}
R.~S. Sutton, D.~Mcallester, S.~Singh, and Y.~Mansour, ``Policy gradient
  methods for reinforcement learning with function approximation,'' in \emph{In
  Advances in Neural Information Processing Systems 12}.\hskip 1em plus 0.5em
  minus 0.4em\relax MIT Press, 2000, pp. 1057--1063.

\bibitem{Atkeson+Santamaria:1997}
\BIBentryALTinterwordspacing
C.~G. Atkeson and J.~C. Santamar\'{i}a, ``A comparison of direct and
  model-based reinforcement learning,'' in \emph{Proceedings of the 1997 {IEEE}
  International Conference on Robotics and Automation}, vol.~4.\hskip 1em plus
  0.5em minus 0.4em\relax {IEEE} Press, 1997, pp. 3557--3564. [Online].
  Available: \url{ftp://ftp.cc.gatech.edu/pub/people/cga/rl-compare.ps.gz}
\BIBentrySTDinterwordspacing

\bibitem{DBLP:journals/ijrr/KoberBP13}
\BIBentryALTinterwordspacing
J.~Kober, J.~A. Bagnell, and J.~Peters, ``Reinforcement learning in robotics:
  {A} survey,'' \emph{I. J. Robotic Res.}, vol.~32, no.~11, pp. 1238--1274,
  2013. [Online]. Available: \url{http://dx.doi.org/10.1177/0278364913495721}
\BIBentrySTDinterwordspacing

\bibitem{Bakker:06icra}
B.~Bakker, V.~Zhumatiy, G.~Gruener, and J.~Schmidhuber, ``Quasi-online
  reinforcement learning for robots,'' in \emph{Proceedings of the 2006 IEEE
  International Conference on Robotics and Automation ICRA}, 2006.

\bibitem{conf/icra/HesterQS10}
\BIBentryALTinterwordspacing
T.~Hester, M.~Quinlan, and P.~Stone, ``Generalized model learning for
  reinforcement learning on a humanoid robot.'' in \emph{ICRA}.\hskip 1em plus
  0.5em minus 0.4em\relax IEEE, 2010, pp. 2369--2374. [Online]. Available:
  \url{http://dblp.uni-trier.de/db/conf/icra/icra2010.html#HesterQS10}
\BIBentrySTDinterwordspacing

\bibitem{Choi:2011:IRL:1953048.2021028}
\BIBentryALTinterwordspacing
J.~Choi and K.-E. Kim, ``Inverse reinforcement learning in partially observable
  environments,'' \emph{J. Mach. Learn. Res.}, vol.~12, pp. 691--730, Jul.
  2011. [Online]. Available:
  \url{http://dl.acm.org/citation.cfm?id=1953048.2021028}
\BIBentrySTDinterwordspacing

\bibitem{Sutton:1998:IRL:551283}
R.~S. Sutton and A.~G. Barto, \emph{Introduction to Reinforcement Learning},
  1st~ed.\hskip 1em plus 0.5em minus 0.4em\relax Cambridge, MA, USA: MIT Press,
  1998.

\bibitem{Abbeel04apprenticeshiplearning}
P.~Abbeel and A.~Y. Ng, ``Apprenticeship learning via inverse reinforcement
  learning,'' in \emph{In Proceedings of the Twenty-first International
  Conference on Machine Learning}.\hskip 1em plus 0.5em minus 0.4em\relax ACM
  Press, 2004.

\bibitem{DBLP:journals/corr/TamarLA16}
\BIBentryALTinterwordspacing
A.~Tamar, S.~Levine, and P.~Abbeel, ``Value iteration networks,'' \emph{CoRR},
  vol. abs/1602.02867, 2016. [Online]. Available:
  \url{http://arxiv.org/abs/1602.02867}
\BIBentrySTDinterwordspacing

\bibitem{Goodfellow-et-al-2016-Book}
\BIBentryALTinterwordspacing
A.~C. Ian~Goodfellow, Yoshua~Bengio, ``Deep learning,'' 2016, book in
  preparation for MIT Press. [Online]. Available:
  \url{http://www.deeplearningbook.org}
\BIBentrySTDinterwordspacing

\bibitem{Lin:1992}
\BIBentryALTinterwordspacing
L.-J. Lin, ``Self-improving reactive agents based on reinforcement learning,
  planning and teaching,'' \emph{Machine Learning}, vol.~8, no. 3--4, pp.
  293--321, 1992. [Online]. Available:
  \url{http://www.cs.ualberta.ca/~sutton/lin-92.pdf}
\BIBentrySTDinterwordspacing

\bibitem{Kearns:2002:NRL:599616.599699}
\BIBentryALTinterwordspacing
M.~Kearns and S.~Singh, ``Near-optimal reinforcement learning in polynomial
  time,'' \emph{Mach. Learn.}, vol.~49, no. 2-3, pp. 209--232, Nov. 2002.
  [Online]. Available: \url{http://dx.doi.org/10.1023/A:1017984413808}
\BIBentrySTDinterwordspacing

\bibitem{DBLP:journals/corr/StadieLA15}
\BIBentryALTinterwordspacing
B.~C. Stadie, S.~Levine, and P.~Abbeel, ``Incentivizing exploration in
  reinforcement learning with deep predictive models,'' \emph{CoRR}, vol.
  abs/1507.00814, 2015. [Online]. Available:
  \url{http://arxiv.org/abs/1507.00814}
\BIBentrySTDinterwordspacing

\bibitem{Bakker02reinforcementlearning}
B.~Bakker, ``Reinforcement learning with long short-term memory,'' in \emph{In
  NIPS}.\hskip 1em plus 0.5em minus 0.4em\relax MIT Press, 2002, pp.
  1475--1482.

\bibitem{DBLP:journals/corr/OhGLLS15}
\BIBentryALTinterwordspacing
J.~Oh, X.~Guo, H.~Lee, R.~L. Lewis, and S.~P. Singh, ``Action-conditional video
  prediction using deep networks in atari games,'' \emph{CoRR}, vol.
  abs/1507.08750, 2015. [Online]. Available:
  \url{http://arxiv.org/abs/1507.08750}
\BIBentrySTDinterwordspacing

\bibitem{DBLP:conf/nips/Atkeson97}
\BIBentryALTinterwordspacing
C.~G. Atkeson, ``Nonparametric model-based reinforcement learning,'' in
  \emph{Advances in Neural Information Processing Systems 10, {[NIPS}
  Conference, Denver, Colorado, USA, 1997]}, 1997, pp. 1008--1014. [Online].
  Available:
  \url{http://papers.nips.cc/paper/1476-nonparametric-model-based-reinforcement-learning}
\BIBentrySTDinterwordspacing

\bibitem{DBLP:journals/corr/Shalev-ShwartzB16}
\BIBentryALTinterwordspacing
S.~Shalev{-}Shwartz, N.~Ben{-}Zrihem, A.~Cohen, and A.~Shashua, ``Long-term
  planning by short-term prediction,'' \emph{CoRR}, vol. abs/1602.01580, 2016.
  [Online]. Available: \url{http://arxiv.org/abs/1602.01580}
\BIBentrySTDinterwordspacing

\end{thebibliography}

\end{document}